\documentclass[11pt]{article}

\usepackage{microtype}
\usepackage{booktabs}
\usepackage{url}
\usepackage{amsmath}
\usepackage{amsthm}
\usepackage[final]{automl}

\usepackage{natbib}

\newcommand{\fundingfooter}{%
  This research has received funding from the Research Council of Lithuania
  (LMTLT), agreement No S-ITP-25-3.}

\makeatletter
\AtBeginDocument{%
  \setlength{\footskip}{0.62in}%
  \fancyfoot[L]{}%
  \fancyfoot[R]{\small\thepage}%
  \fancypagestyle{plain}{%
    \fancyhf{}%
    \fancyfoot[L]{%
      \parbox{\textwidth}{%
        \small\c@nferencestring
        \hfill
        \small\textcopyright\,\@utomlyear\ the authors, released under
        \href{https://creativecommons.org/licenses/by/4.0/}{CC BY 4.0}\\[3pt]
        \scriptsize\fundingfooter}}%
  }%
}
\makeatother

\title{Evolutionary Architecture Search for Chlorophyll-$a$ Prediction in Lakes using Sentinel-2}

\author[1, 2, 3]{\nameemail{K\"ur\c{s}at K\"om\"urc\"u}{kursat.komurcu@mif.vu.lt}}
\author[1]{\nameemail{Linas Petkevi\v{c}ius}{linas.petkevicius@mif.vu.lt}}

\affil[1]{Vilnius University, Institute of Computer Science, Artificial Intelligence Methods Lab }
\affil[2]{IRISA, Universite Bretagne Sud}
\affil[3]{European Commission Joint Research Center}

\hypersetup{%
  pdfauthor={Kursat Komurcu, Linas Petkevicius},
  pdftitle={Evolutionary Architecture Search for Small Tabular Earth
    Observation Problems},
  pdfsubject={Neural architecture search for Earth observation},
  pdfkeywords={neural architecture search, regularized evolution, Earth
    observation, Sentinel-2, tabular data, onboard AI}
}

\begin{document}

\maketitle

\begin{abstract}
  Small tabular datasets with expert-designed spectral features are the norm in
  operational Earth observation, and the networks applied to them are typically
  hand-designed. We revisit one such published model -- a Sentinel-2 algal
  bloom classifier -- and ask what architecture search adds, holding the task,
  the features and the lake-level train/test split of the original study fixed.
  Searching an extended multilayer-perceptron space with regularized evolution,
  and selecting on inner-cross-validation AUC only, we find networks that
  improve held-out AUC from 0.790 to 0.820 and accuracy from 0.733 to 0.748
  while using 409 trainable parameters, 26 times fewer than the strongest
  hand-designed reference. The search converges on a consistent recipe -- a
  single narrow layer, RMS normalisation, $\tanh$ activation, step-decayed
  RMSprop and weight averaging -- that a practitioner would be unlikely to
  reach by default. At 1.6\,kB the resulting model is small enough to serve as
  an onboard screening trigger, which is the setting that motivates the work. 
  Code: \url{https://github.com/VU-AIML/automl4eo-bloom-nas}.
\end{abstract}

\section{Introduction}

Operational Earth observation tasks are frequently small and tabular: in-situ
labels are expensive, so datasets run to a few thousand rows of expert-designed
spectral features rather than to millions of images. The networks applied to
them are correspondingly hand-designed, as remains common for tabular problems
generally \citep{gorishniy2021,grinsztajn2022}. Neural architecture search is a
mature toolbox \citep{elsken2019}, but is applied overwhelmingly to large-scale
vision; its behaviour in this regime is much less characterised, and it is not
obvious that a search should help at all when the task offers so little signal
to search against.

We study it on one such task. \citet{grendaite2024} assembled a dataset of
Baltic lakes matching in-situ observations to Sentinel-2 overpasses and trained
a family of hand-designed networks -- plain stacks, residual stacks, and stacks
with parallel dense heads -- on per-lake spectral features, reporting 71.6\,\%
accuracy for bloom classification. Holding their task, features and lake-level
split fixed, we ask what architecture search adds.

The question matters beyond accuracy. Downlink bandwidth rather than sensing is
the binding constraint for many Earth observation products, and onboard
inference has been demonstrated in orbit as a way to transmit only informative
data \citep{giuffrida2020,giuffrida2022,marin2021}. How small a competitive
model can be is therefore an operational question, not only an efficiency one.

We search an extended multilayer-perceptron (MLP) space with regularized
evolution, selecting on inner-cross-validation AUC alone. Our contributions
are:

\begin{itemize}
\item Searched architectures that outperform all three hand-designed reference
  networks on held-out AUC, accuracy and $F_1$, with all ten highest-ranked
  configurations beating every baseline on AUC.
\item A 409-parameter model -- 26 to 47 times smaller than the references --
  that achieves this, making onboard screening plausible at 1.6\,kB.
\item A characterisation of the design region the search converges on, and an
  account of why selection must be driven by AUC at this sample size.
\end{itemize}

\section{Task, data and protocol}

\paragraph{Data}
The dataset released with \citet{grendaite2024} contains 1151 observations from
429 monitoring sites in Lithuania, Latvia and Estonia, collected between 2017
and 2021.\footnote{The public release appears to predate the revision
  referenced by the accompanying preprocessing script, so our baseline numbers
  are a reimplementation on the public data rather than a bit-exact
  reproduction of the published figures.} Each row holds Sentinel-2 surface
reflectances together with derived radiometric indices of the kind established
for chlorophyll retrieval in turbid inland waters \citep{mishra2012}, for one
lake on one date, with a binary bloom label. Classes are close to balanced
(624 non-bloom, 527 bloom), so accuracy is interpretable against a 52.9\,\%
majority-class floor. We use the same 14 features as the reference study.

\paragraph{Split}
We reproduce the reference split exactly, including its random seed: 70\,\% of
the unique site codes are drawn for training and every measurement from a
selected lake follows, giving 801 training and 350 test rows over 129 held-out
lakes. Grouping by lake rather than by row is essential here, since a single
lake contributes up to 17 observations and row-level splitting would allow a
model to memorise individual water bodies.

\paragraph{Nested evaluation}
All search signal comes from a lake-grouped 5-fold cross-validation
\emph{inside} the 801-row training split; the 350-row test split is opened
once, after the search has terminated. This nesting is not a formality at this
scale: selecting the best of thousands of candidates on the same data used to
report performance yields the maximum of a noisy statistic rather than a
generalisation estimate, and the resulting optimism can rival the differences
between the methods compared \citep{varma2006,cawley2010}.

\paragraph{Selection metric}
We select on AUC rather than accuracy, for two reasons. Operationally, an
onboard trigger ranks candidate scenes against a transmission budget, so the
decision threshold is a deployment parameter rather than a property of the
model. Empirically, accuracy on a 160-row validation fold is dominated by how a
handful of near-threshold observations happen to fall, whereas AUC uses the
full ordering: the seed-to-seed standard deviation of held-out accuracy reaches
0.019 in our runs while that of AUC stays at or below 0.008. Since the variance
of the selection criterion, not only its bias, governs how badly model
selection overfits \citep{cawley2010}, this choice does more work than the
metric definition alone suggests.

\section{Search space and method}

\paragraph{Space}
The space (Table~\ref{tab:space}) stays within the MLP family, so that the
comparison against the hand-designed networks is like-for-like, but widens the
axes those networks left implicit; it contains all three reference
architectures as individual points. Five block types are available: plain dense
blocks, residual blocks, the parallel-dense ``multi-head'' block used in the
reference implementation, gated linear units, and gated residual blocks, the
last two following components that have proved useful in tabular deep learning
\citep{gorishniy2021}. Width is specified by a base width together with a
profile, since the reference networks use tapering stacks that a single global
width cannot express. Optimiser, schedule, warmup, label smoothing, class
weighting, gradient clipping and weight averaging \citep{izmailov2018} are
searched alongside the topology. Candidates above 200k trainable parameters are
rejected: with 801 training rows, larger networks cannot be constrained by the
data, and the onboard setting rules them out regardless.

\begin{table}[t]
  \centering
  \footnotesize
  \caption{Search space, approximately $10^{12}$ configurations before the
    parameter-budget constraint.}
  \label{tab:space}
  \begin{tabular}{llll}
    \toprule
    Dimension & Values & Dimension & Values \\
    \midrule
    Blocks        & 1--6                       & Optimiser       & AdamW, Adam, SGD, RMSprop \\
    Base width    & 16--256 (9 values)         & Learning rate   & $10^{-4}$--$3\times10^{-2}$ \\
    Width profile & constant, taper,           & Schedule        & constant, cosine, step \\
                  & \quad expand, hourglass    & Warmup fraction & 0.0, 0.05, 0.1 \\
    Block type    & plain, residual,           & Weight decay    & 0--$10^{-2}$ \\
                  & \quad multi-head, GLU, GRN & Batch size      & 8--256 \\
    Heads         & 2, 4, 8                    & Epochs          & 50--600 \\
    Normalisation & batch, layer, RMS, none    & Label smoothing & 0.0--0.15 \\
    Activation    & ReLU, GELU, ELU,           & Class weighting & off, on \\
                  & \quad SiLU, Mish, $\tanh$  & Gradient clip   & off, 0.5, 1.0, 5.0 \\
    Dropout       & 0.0--0.5                   & Weight EMA      & off, 0.99, 0.995, 0.999 \\
    Input dropout & 0.0--0.3                   &                 & \\
    \bottomrule
  \end{tabular}
\end{table}

\paragraph{Method}
Random search is a strong default in spaces of this kind
\citep{bergstra2012,li2019} and we used it in preliminary work, but the space
above is roughly a trillion configurations and rewards local exploitation. We
therefore use regularized evolution \citep{real2019}: an initial random
population of 64, then rounds in which 20 children are bred by tournament
selection (size 10) and one to three random dimension edits, with the oldest
individuals retired regardless of score. Age-based rather than score-based
retirement is the important detail at this scale: a single cross-validation
estimate carries one to three points of standard deviation, so a configuration
can enter the population through luck, and score-based retirement would let it
persist indefinitely while its mutations took over. Retirement by age forces a
lineage to re-earn its score at every generation.

Each candidate is scored by 5 folds $\times$ 3 weight seeds, averaged. The run
evaluated 2720 candidates (1827 distinct architectures after collapsing
configurations that differ only in inactive dimensions) in approximately five
hours on 20 CPU cores. The networks are small enough that CPU evaluation
parallelised across trials is faster than GPU execution.

\section{Results}

Table~\ref{tab:main} reports the comparison. To ensure that any gain is
attributable to the searched configuration rather than to an incidentally
better training loop, the reference architectures are expressed in the same
space and trained by the same code, using the optimiser and schedule settings
of the original implementation.

\begin{table}[t]
  \centering
  \small
  \caption{Held-out performance, mean $\pm$ standard deviation over 10 weight
    seeds. Selection used inner-CV AUC only. NAS-1/6/8 are the three
    highest-ranked distinct architectures; the remaining seven of the top ten
    fall within 0.002 AUC of these.}
  \label{tab:main}
  \begin{tabular}{lccccr}
    \toprule
    Model & CV AUC & Test AUC & Test acc. & $F_1$ & Params \\
    \midrule
    Baseline-MLP & 0.8055 & $0.7902 \pm .006$ & $0.7306 \pm .009$ & 0.7241 & 10\,625 \\
    Baseline-Res & 0.7588 & $0.7543 \pm .010$ & $0.6989 \pm .010$ & 0.6746 & 32\,129 \\
    Baseline-MHA & 0.7914 & $0.7785 \pm .008$ & $0.7329 \pm .011$ & 0.7266 & 19\,073 \\
    \midrule
    NAS-1 & 0.8240 & $0.8181 \pm .001$ & $0.7460 \pm .004$ & 0.7442 & \textbf{409} \\
    NAS-6 & 0.8238 & $\mathbf{0.8203} \pm .001$ & $0.7449 \pm .003$ & 0.7449 & 433 \\
    NAS-8 & 0.8237 & $0.8187 \pm .001$ & $\mathbf{0.7483} \pm .003$ & \textbf{0.7466} & \textbf{409} \\
    \bottomrule
  \end{tabular}
\end{table}

The searched models improve held-out AUC by 3.0 points over the best baseline
and accuracy by 1.5 points, using 409 parameters against 10\,625 for the
strongest reference network -- a 26-fold reduction, or 47-fold against the
multi-head variant. All ten highest-ranked configurations beat every baseline
on AUC, so the result does not rest on a single fortunate candidate.

The searched models are also markedly more stable, with a seed-to-seed standard
deviation of 0.001 in AUC against 0.006--0.010 for the baselines, which we
attribute to the weight averaging the search selected \citep{izmailov2018}. The
optimism of the cross-validation estimate -- the gap between inner-CV and
held-out AUC -- is only 0.006 for the selected models, indicating that
selection on AUC at this budget is not materially inflated by the selection
bias \citet{cawley2010} describe.

\paragraph{What the search converged on}
Among the 100 highest-ranked configurations, several choices are unanimous:
plain dense blocks, RMSprop, step decay, RMS normalisation, $\tanh$ activation,
weight EMA with decay 0.999, and class weighting. The median configuration uses
a single layer of width 24 trained for 300 epochs. Three of these sit away from
what a practitioner would pick by default. The $\tanh$ activation has largely
been displaced by ReLU-family functions, yet it is bounded, which limits how
confidently a small network can commit on the 801 training rows available.
Weight averaging plays a similar role and is the main source of the stability
noted above. Most strikingly, the search settles on a single hidden layer: the
depth of the reference networks is not merely unnecessary but actively
unhelpful at this sample size. We did not anticipate this combination, which is
the argument for searching rather than designing.

\section{Discussion: onboard feasibility}

At 409 parameters the selected model occupies 1.6\,kB in float32 and requires
on the order of 400 multiply-accumulates per lake. On the class of processor
flown in onboard-AI demonstrations \citep{giuffrida2022} this cost is
negligible, and the model could screen candidate observations in orbit so that
only lakes flagged as blooming consume downlink capacity -- the same argument
that motivated onboard cloud rejection \citep{giuffrida2020}.

We should be precise about what this does and does not establish. The features
used here derive from atmospherically corrected Level-2A products computed on
the ground, together with water masking; porting that chain to the satellite is
a separate and harder problem \citep{marin2021}. What our result establishes is
that the \emph{decision layer} of such a pipeline is essentially free, and that
the capacity a hand-designed model spends on it can be reduced by more than an
order of magnitude at no cost -- indeed with a gain.

\section{Limitations and future work}

The evaluation rests on one dataset and one train/test split. We preserved the
reference split deliberately, for comparability, but a single 350-row test
split carries roughly 2.4 points of standard error on accuracy, and we have not
measured how the ranking behaves across resampled lake splits. The public data
release also appears to predate the revision used for the published numbers, so
our baselines are a reimplementation rather than an exact reproduction. The
search itself converged narrowly: the ten highest-ranked configurations are
minor variations of one architecture, and duplicate suppression with periodic
random re-injection would characterise the space more broadly.

Three extensions follow naturally: the same protocol applied to the
chlorophyll-$a$ regression task released with the reference study; feature
selection as a search dimension, which is operationally relevant since reading
fewer bands means less onboard processing; and quantisation-aware search, which
would turn the onboard claim from a parameter count into a measured latency and
energy figure on representative hardware.

\bibliography{refs}

\end{document}